\documentclass{article}
\usepackage{spconf,amsmath,bm,comment,cite}
\usepackage[pdftex]{graphicx} 

\newcommand{\argmax}{\mathop{\rm arg~max}\limits}

\title{PDH : Probabilistic deep hashing \\based on MAP estimation of Hamming distance}
\twoauthors
  {Yosuke Kaga, Masakazu Fujio, Kenta Takahashi}
	{Hitachi, Ltd.}
  {Tetsushi Ohki, Masakatsu Nishigaki}
	{Shizuoka University}
\begin{document}

\begin{minipage}[t]{160mm}

\section*{IEEE Copyright Notice}
\large
© 2019 IEEE. Personal use of this material is permitted. Permission from IEEE must be obtained for all other uses, in any current or future media, including reprinting/republishing this material for advertising or promotional purposes, creating new collective works, for resale or redistribution to servers or lists, or reuse of any copyrighted component of this work in other works.\\

Accepted by the 26th IEEE International Conference on Image Processing(ICIP2019)
\end{minipage}
\newpage

%\ninept
%
%\renewcommand{\baselinestretch}{0.90}
\def\ninept{\def\baselinestretch{.05}\let\normalsize\small\normalsize}
\maketitle
\begin{abstract}
With the growth of image on the web,  research on hashing which enables high-speed image retrieval has been actively studied.
In recent years, various hashing methods based on deep neural networks have been proposed and achieved higher precision than the other hashing methods.
In these methods, multiple losses for hash codes and the parameters of neural networks are defined.
They generate hash codes that minimize the weighted sum of the losses.
Therefore, an expert has to tune the weights for the losses heuristically, and the probabilistic optimality of the loss function cannot be explained.
In order to generate explainable hash codes without weight tuning, we theoretically derive a single loss function with no hyperparameters for the hash code from the probability distribution of the images.
By generating hash codes that minimize this loss function, highly accurate image retrieval with probabilistic optimality is performed.
We evaluate the performance of hashing using MNIST, CIFAR-10, SVHN and show that the proposed method outperforms the state-of-the-art hashing methods.
%about 100 to 150words
\end{abstract}
\begin{keywords}
hashing, image retrieval, deep learning, binary representation
\end{keywords}
\vspace{-0.5pc}
\section{Introduction}
\label{sec:intro}
\vspace{-0.5pc}
In recent years, an enormous amount of images has been managed on the web.
In order to search images from such big data, search algorithms for content-based image retrieval (CBIR) have been studied.
The problem of finding the data closest to the query on a certain distance measure is called the nearest neighbor (NN) search.
In order to perform this NN search as it is, a huge amount of calculation time is required, so fast methods for obtaining an approximate nearest neighbor (ANN) has been studied.

Hashing is a method which is actively studied for ANN search \cite{Wang18}.
In the hashing, images are mapped to binary hash codes while preserving semantic similarities between images.
By using the hash codes, the retrieval for a huge amount of images can be performed with small computational time.
Many hashing methods have been proposed \cite{Datar04,Kulis09,Yair09, Gong13, Brian09,Liu12,Shen15,Rongkai14,Liu16,Yang18,Hanjiang15,Masoumeh18}.
Especially the supervised hashing methods learn hash functions by using supervised information such as image classes, thus high accuracy can be achieved for specific datasets \cite{Liu12,Shen15}.
Furthermore, deep learning is applied to the supervised hashing, and higher accuracy is achieved than other types of hashing methods\cite{Rongkai14,Liu16,Yang18,Hanjiang15,Masoumeh18}.
In these methods, multiple losses for hash codes and neural network parameters are defined.
They generate hash codes that minimize the weighted sum of the losses.
Therefore, an expert has to tune the weights for the losses heuristically, and the probabilistic optimality of the loss function cannot be explained.

In this paper, we theoretically derive a single loss function with no hyperparameters for the hash codes from the probability distribution of the images.
We obtain convolutional neural networks that minimize the loss function.
By binarizing the output of that networks, hash codes for image retrieval are calculated.
We show that the Hamming distance between binarized hash codes is equivalent to the MAP estimation of the ideal hash distance.
Based on this probabilistically optimal hash codes, we can perform highly accurate image retrieval.
Our contributions in this paper are as follows: (1) We introduce new loss function with no hyperparameters for hash codes and show that the Hamming distance between hash codes has probabilistic optimality. (2) We show the experimental results that our hashing method outperforms the other state-of-the-art hashing methods in image retrieval.
\begin{figure*}[t]
\begin{minipage}{5cm}
  \centering
  \centerline{\includegraphics[width=5cm]{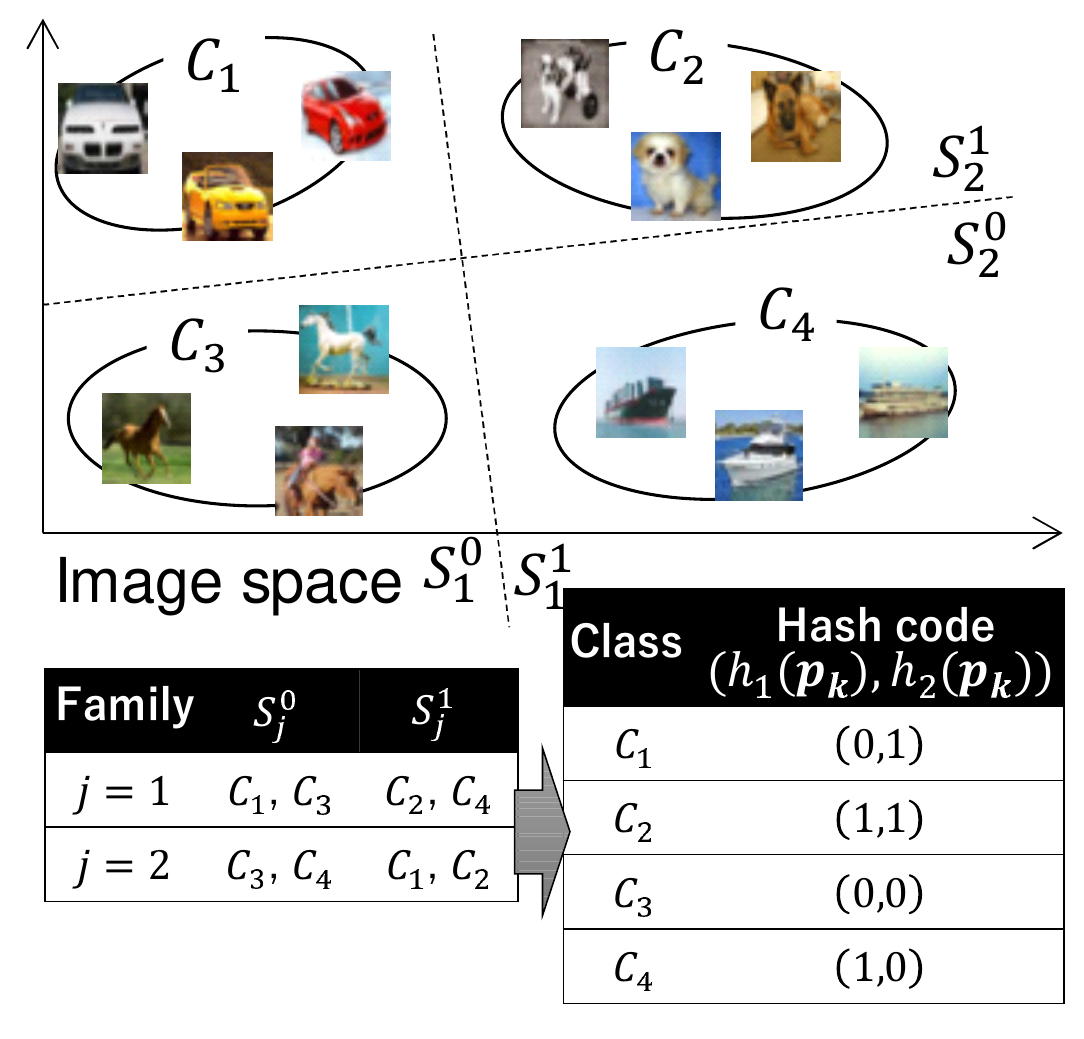}}
\caption{An example of image classes and hash codes.}
\label{fig:class}
\end{minipage}
\begin{minipage}{13cm}
  \centering
  \centerline{\includegraphics[width=12cm]{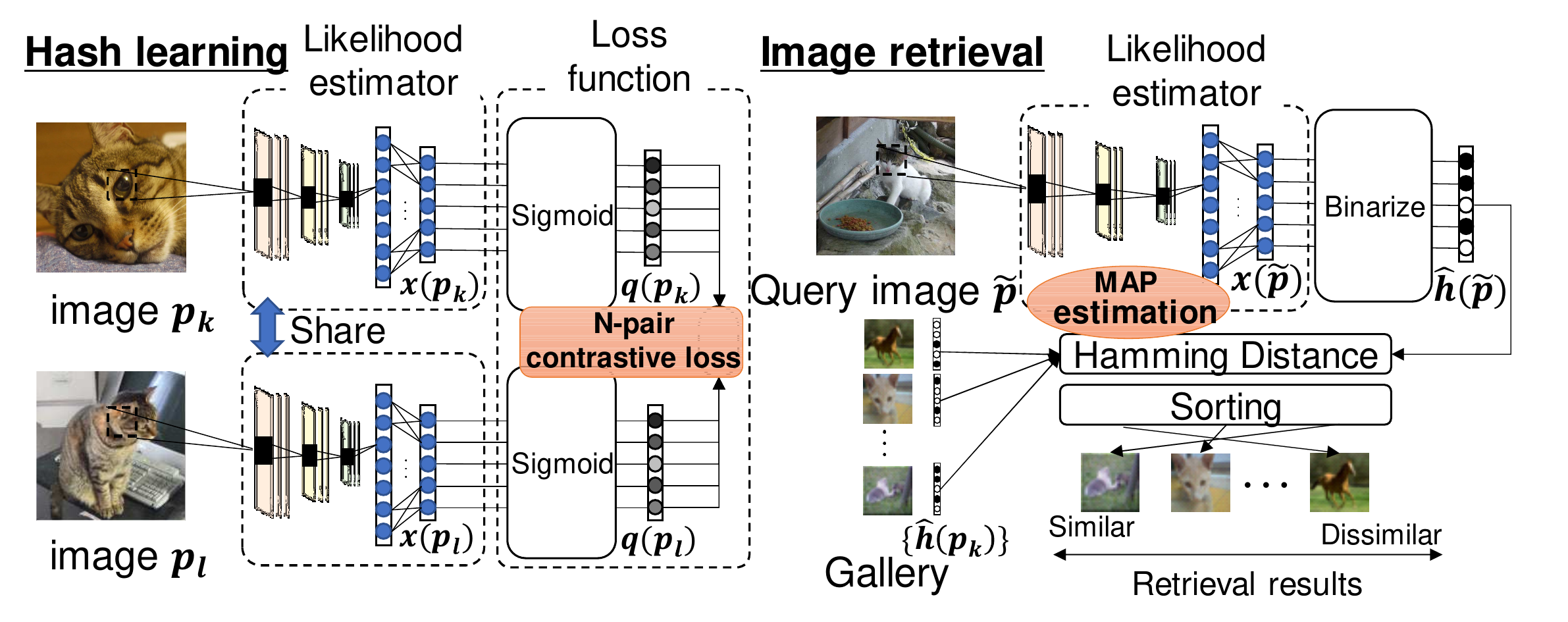}}
  \vspace{-0.5cm}
\caption{The architecture of our probabilistic deep hashing.}
\label{fig:architecture}
\end{minipage}
\end{figure*}
\vspace{-0.5pc}
\section{Problem definition}
\label{sec:problem}
\vspace{-0.5pc}
We define the problem for generating hash codes that enable image retrieval with high precision.
In this paper, we derive an ideal hash function from image probabilistic distributions.
Let the retrieval target images be $\bm{p} = \{\bm{p_k}\}_{k = 1}^{N_p} \in \mathcal{R}^{m}$ and its class set be $\bm{C} = \{C_i\}_{i=1}^{N_c}$, where $N_p$ is the number of the retrieval target images,$m$ is the dimension of an image and $N_c$ is the number of the class, respectively.
Fig.\ref{fig:class} shows an example of image classes $C_i$ on the image space.
We assume that every image belongs to a single image class.
Let the event that an image $\bm{p_k}$ belongs to class $C_i$ denote $\bm{p_k} \gets C_i$.

Next, let families of image class set be $\bm{S} =$ $\{(\bm{S_j^0},$ $\bm{S_j^1})\}_{j=1}^{n}$, where $n$ is the number of families.
The families satisfy $\{\bm{S_j^0},\bm{S_j^1}\} = \bm{C}$ and $\bm{S_j^0} \cap \bm{S_j^1} = \emptyset$.
Therefore, $(\bm{S_j^0},\bm{S_j^1})$ divides image classes into two subsets.
The examples of these families are shown in Fig.\ref{fig:class}.

By using the image class set $\bm{C}$ and the families of image classes $\bm{S}$, we define ideal hash functions $h = \{h_j\}_{j=1}^{n}$, \\
$h_j : \mathcal{R}^m \to \{0,1\}$ as follows:
\vspace{-0.5pc}
\begin{eqnarray}
h_j(\bm{p_k}) =
\begin{cases}
\scalebox{0.9}{$\displaystyle 1 $} & \scalebox{0.9}{$\displaystyle \text{if $C_i \in S_j^1$ where $\bm{p_k} \gets C_i$} .$} \\
\scalebox{0.9}{$\displaystyle 0 $} & \scalebox{0.9}{$\displaystyle \text{if $C_i \in S_j^0$ where $\bm{p_k} \gets C_i$} .$} 
\end{cases}
\end{eqnarray}
When image $\bm{p_k}$ belongs to class $C_i$,
the value of hash function $h_j(\bm{p_k})$ depends on the families $\bm{S_j^0}$,$\bm{S_j^1}$ that include class $C_i$.

Based on these hash functions, we define the mapping from images to hash codes; $h(\bm{p_k}) = (h_1(\bm{p_k}), h_2(\bm{p_k}),$ $\cdots,$ $h_n(\bm{p_k}))$.
In the image retrieval based on hashing, the Hamming distances between the hash codes are used as image dissimilarities.
When $\bm{p_k} \gets C_{i_1}$ and $\bm{p_l} \gets C_{i_2}$, the Hamming distance between hash codes is calculated as follows:
\vspace{-0.5pc}
\begin{eqnarray}
& \scalebox{0.9}{$\displaystyle d(h(\bm{p_k}),h(\bm{p_l})) = \sum\nolimits_{j=1}^{n} | h_j(\bm{p_k}) - h_j(\bm{p_l}) | $} \nonumber \\
&\scalebox{0.9}{$\displaystyle = \begin{cases}
0 & \text{if } i_1 = i_2. \\
\sum_{j=1}^{n} t_j(\bm{p_k},\bm{p_l}) & \text{otherwise.}
\end{cases} $} \\
& \scalebox{0.9}{$\displaystyle t_j(\bm{p_k},\bm{p_l}) = \begin{cases}
0 & \text{if  } ( C_{i_1},  C_{i_2} \in S_j^0 ) \lor ( C_{i_1}, C_{i_2} \in S_j^1 ) .\\
1 & \text{otherwise.}
\end{cases} $}
\end{eqnarray}
In order to minimize the probability of hash collision for images in different classes, the families of image class set $\bm{S}$ need to satisfy the following properties: (1) Each image class $C_i$ are randomly assigned to $\bm{S_j^0}$ or $\bm{S_j^1}$ independently. (2) $\forall i,j : Pr(C_i \in S_j^0) = Pr(C_i \in S_j^1) = 0.5$.
Based on this assumption, the expected value of Hamming distance is calculated as follows:
\vspace{-0.5pc}
\begin{eqnarray}
&\scalebox{0.9}{$\displaystyle E_{\bm{p_k},\bm{p_l}} \left[ d(h(\bm{p_k}),h(\bm{p_l})) \right] = 
\begin{cases}
0 & \text{if  } i_1 = i_2. \\
n/2 & \text{otherwise.}
\end{cases} $} \label{eq:epp}
\end{eqnarray}
This is the requirement for ideal hash distance under our assumption.
This ideal hash distance is small when input images belong to the same class and large when the images belong to a different class.
According to this property, image retrieval based on the hash function would achieve high precision.
The purpose of this paper is to find hash functions that Hamming distances between the hash codes meet the above requirement.
\begin{comment}
\begin{figure*}[t]
  \centering
  \centerline{\includegraphics[width=18cm]{architecture}}
\caption{The architecture of our probabilistic deep hashing.}
\label{fig:architecture}
%
\end{figure*}
\end{comment}
\vspace{-0.5pc}
\section{Probabilistic deep Hashing}
\label{sec:proposed}
\vspace{-0.5pc}
In accordance with the requirement for ideal hash functions in section \ref{sec:problem}, we propose probabilistic deep hashing (PDH) which can perform image retrieval with probabilistic optimality.
\vspace{-0.5pc}
\subsection{The concept of the PDH}
\label{subsec:concept}
\vspace{-0.5pc}
First, we show the concept of the proposed method.
Our PDH has two characteristics as follows:
\begin{enumerate}
\item We derive the property that the ideal hash codes should satisfy from the probability distribution of the image.
We perform pairwise learning to bring the expected value of the Hamming distance between the hash codes closer to the ideal hash distance as described in Sec.\ref{subsec:hash}.
This allows us to learn the hash functions that have close property to ideal hash functions.
\item We approximate the hash functions with convolutional neural networks (CNN). 
By assuming the output of the CNN as posterior probabilities of the hash function outputs, we show that the Hamming distance between the obtained hash codes is equal to the MAP estimation of the ideal hash distance as described in Sec.\ref{subsec:retrieval}.
As a result, image retrieval with probabilistic optimality can be realized with high-speed bit operation.
\end{enumerate}
By utilizing these characteristics, the PDH can perform image retrieval with high precision.
\vspace{-0.5pc}
\subsection{PDH Architecture}
\label{subsec:architecture}
\vspace{-0.5pc}
We show the architecture of the proposed method in Fig.\ref{fig:architecture}.
The method consists of two parts: hash learning and image retrieval.
In the hash learning part, we learn a likelihood estimator based on new loss function, and the detail of the hash learning is described in Sec.\ref{subsec:hash}.
In the image retrieval part, hash codes of images are obtained by the likelihood estimator, and ANN search is performed based on Hamming distance between hash codes.
The detail of the image retrieval is described in Sec.\ref{subsec:retrieval}.
\vspace{-0.5pc}
\subsection{Hash learning}
\label{subsec:hash}
\vspace{-0.5pc}
\begin{comment}
\vspace{-0.5pc}
\begin{eqnarray}
h_j(\bm{p}) =
\begin{cases}
\scalebox{0.9}{$\displaystyle 1 $} & \scalebox{0.9}{$\displaystyle \text{if $S_j(C_i) = 1$ where $\bm{p} \gets D_i$} $} \\
\scalebox{0.9}{$\displaystyle 0 $} & \scalebox{0.9}{$\displaystyle \text{otherwise} $}
\end{cases}
\end{eqnarray}
\end{comment}
First, we define the prior probability of image class $C_i$ and the likelihood for image $\bm{p_k}$ when $\bm{p_k}$ belongs to image class $C_i$ as follows:
\vspace{-0.5pc}
\begin{eqnarray}
\scalebox{0.9}{$\displaystyle \forall i : Pr(\bm{p_k} \gets C_i) = 1 / N_c .$} \\
\scalebox{0.9}{$\displaystyle D_i(\bm{p_k}) := Pr(\bm{p_k} | \bm{p_k} \gets C_i) .$}
\end{eqnarray}
Here we derive the posterior probability $q_j (\bm{p_k})$ of $h_j(\bm{p_k})=1$ when $\bm{p_k}$ is observed based on the bayes' theorem.
\vspace{-0.5pc}
\begin{eqnarray}
q_j(\bm{p_k}) :=& Pr(h_j(\bm{p_k}) = 1|\bm{p_k}) \nonumber \\
=& \frac{Pr(\bm{p_k}|h_j(\bm{p_k})=1) Pr(h_j(\bm{p_k})=1)}{Pr(\bm{p_k})} \nonumber \\
=&\frac{Pr(\bm{p_k}|h_j(\bm{p_k})=1) Pr(h_j(\bm{p_k})=1)}{\sum_{g = 0}^{1} Pr(\bm{p_k}|h_j(\bm{p_k})=g) Pr(h_j(\bm{p_k})=g)} \nonumber \\
=& \frac{1}{1 + \frac{Pr(\bm{p_k}|h_j(\bm{p_k})=0) Pr(h_j(\bm{p_k})=0)}{Pr(\bm{p_k}|h_j(\bm{p_k})=1) Pr(h_j(\bm{p_k})=1)}} \nonumber \\
=& \frac{1}{1 + \frac{\sum_{v=1}^{N_c} Pr(\bm{p_k}|\bm{p_k} \gets C_v) Pr(\bm{p_k} \gets C_v|h_j(\bm{p_k}) = 0)}{\sum_{w=1}^{N_c} Pr(\bm{p_k}|\bm{p_k} \gets C_w) Pr(\bm{p_k} \gets C_w|h_j(\bm{p_k}) = 1)}} .
\end{eqnarray}
Here, $Pr(\bm{p_k} \gets C_v | h_j(\bm{p_k}) = u)$ is calculated as follows:
\vspace{-0.5pc}
\begin{eqnarray}
& \scalebox{0.85}{$\displaystyle Pr(\bm{p_k} \gets C_v | h_j(\bm{p_k}) = u) = 
\begin{cases}
1 / \alpha_j^1 & \text{if } u = 1 \land C_v \in \bm{S_j^1} .\\
1 / \alpha_j^0 & \text{if }  u = 0 \land C_v \in \bm{S_j^0} .\nonumber \\
0 & \text{otherwise.}
\end{cases} $} \\
& \alpha_j^1 = |\bm{S_j^1}|, \alpha_j^0 = |\bm{S_j^0}| ,
\end{eqnarray}
where $|\bm{S_j^u}|$ is the number of image classes that $C_i \in \bm{S_j^u}$.
Also, let a flag value for image class be $t_{ij} = 1$ when $C_i \in S_j^1$ and $t_{ij} = 0$ otherwise.
Finally, $q_j(\bm{p_k})$ is calculated as follows:
\vspace{-0.5pc}
\begin{eqnarray}
q_j(\bm{p_k}) =& \frac{1}{1 + \exp(x_j(\bm{p_k}))} .\\
x_j(\bm{p_k}) = & \log \frac{\alpha_j^1 \sum_{v=1}^{N_c} D_v(\bm{p_k}) (1-t_{vj})}{\alpha_j^0 \sum_{w=1}^{N_c} D_w(\bm{p_k}) t_{wj}} .
\end{eqnarray}
Here, $q_j(\bm{p_k})$ is obtained as the output of the sigmoid function, and $x_j(\bm {p_k})$ is the log-likelihood ratio for $h_j (\bm {p_k}) = 1$ and $h_j (\bm {p_k}) = 0$ when image $\bm{p_k}$ is observed.
Then, the log-likelihood ratio takes a large negative value when the probability that $h_j(\bm {p_k}) = 0 $ is high, and it takes a large positive value when the probability that $h_j(\bm{p_k}) = 1 $ is high.
If the likelihood for images $D_i(p_k)$ is known,  $x_j(\bm {p_k})$ can be calculated.
However, it is difficult to obtain $D_i(\bm{p_k})$ in advance, and we cannot calculate $x_j(\bm{p_k})$ directly.
In the proposed method, the log-likelihood ratio $\bm{x}(\bm{p_k}) = \{x_j(\bm{p_k}) \}_ {j = 1} ^ {n}$ is approximated by convolutional neural networks as shown in Fig.\ref{fig:architecture}.

Next, we derive the Hamming distance between hash codes by using $q_j(\bm{p_k})$. 
With the posterior probabilities $q_j(\bm{p_k})$, the distance between hash codes can also be represented by a probability distribution.
Assuming that the output of the hash functions $\{h_j(\bm{p_k})\}$ is independent respectively, we calculate the expected value of the Hamming distance $E_{\bm{p_k},\bm{p_l}}[d(h(\bm{p_k}),h(\bm{p_l}))]$ as follows:
\vspace{-0.5pc}
\begin{eqnarray}
&\scalebox{0.80}{$\displaystyle E_{\bm{p_k},\bm{p_l}}\left[d(h(\bm{p_k}),h(\bm{p_l}))\right] $} \nonumber \\
=& \scalebox{0.80}{$\displaystyle E_{\bm{p_k},\bm{p_l}}\left[ \sum\nolimits_{j=1}^{n} |h_j(\bm{p_k}) - h_j(\bm{p_l})|\right] $} \nonumber \\
=& \scalebox{0.80}{$\displaystyle \sum\nolimits_{j=1}^{n} E_{\bm{p_k},\bm{p_l}}\left[ \left|h_j(\bm{p_k}) - h_j(\bm{p_l})\right| \right]  $} \nonumber \\
=& \scalebox{0.80}{$\displaystyle \sum\nolimits_{j=1}^{n} Pr(|h_j(\bm{p_k}) - h_j(\bm{p_l})| = 1|\bm{p_k}, \bm{p_l}) $} \nonumber \\
=& \scalebox{0.80}{$\displaystyle \sum\nolimits_{j=1}^{n} \left\{ q_j(\bm{p_k}) (1-q_j(\bm{p_l})) + (1-q_j(\bm{p_k})) q_j(\bm{p_l}) \right\} .$} \label{eq:edpp}
\end{eqnarray}
By using this expected value of Hamming distance, we define a loss function for the hash codes.
The ideal values of the expected value of Hamming distance are shown in the Eq.(\ref{eq:epp}).
We define a $L_2$ loss function between $E_{\bm{p_k},\bm{p_l}}[d(h(\bm{p_k}),h(\bm{p_l}))]$ and the ideal values.
As a mini-batch of a training set for the loss function, $\bm{\bar{p}} = \{(\bm{\bar{p}_i},\bm{\bar{p}_i^{\prime}})\}_{i = 1}^{N_c}$ where $\bm{\bar{p}_i}, \bm{\bar{p}_i^{\prime}} \gets C_i$ is sampled from $\bm{p}$.
Then, the loss function is calculated as follows:
\vspace{-1pc}
\begin{eqnarray}
&\scalebox{0.80}{$\displaystyle L(\bm{\bar{p}}) = \sum\nolimits_{i=1}^{N_c} \left\{ \left\{E_{\bm{\bar{p}_i},\bm{\bar{p}_i^{\prime}}}\left[d(h(\bm{\bar{p}_i}),h(\bm{\bar{p}_i}^{\prime}))\right]\right\}^{2} +  \right. $} \nonumber \\
&\scalebox{0.80}{$\displaystyle \left. \sum\nolimits_{r=1, r \neq i}^{N_c} \max\left\{\frac{n}{2} - E_{\bm{\bar{p}_i},\bm{\bar{p}_r^{\prime}}}\left[d(h(\bm{\bar{p}_i}),h(\bm{\bar{p}_r}^{\prime}))\right], 0\right\}^{2} \right\} . $} \label{eq:Lhat}
\end{eqnarray}
This loss function is N-pair contrastive loss, which was obtained by modifying contrastive loss \cite{Chopra05} by being inspired by N-pair loss \cite{Kihyuk16}.
We sample randomly mini-batch images $\bm{\bar{p}}$ from $\bm{p}$, input them to N-pair contrastive loss function Eq.(\ref{eq:Lhat}), Eq.(\ref{eq:edpp}), and update the parameters of CNN so that the loss becomes small.
By repeating this pairwise learning process, hash functions $h$ that meet the requirements for ideal hash functions can be obtained.
\begin{table*}[t]
\footnotesize
  \begin{center}
    \caption{Mean average precision of hashing methods.}
    \begin{tabular}{l|p{8mm}p{8mm}p{8mm}p{8mm}|p{8mm}p{8mm}p{8mm}p{8mm}|p{8mm}p{8mm}p{8mm}p{8mm}} \hline
      Method & \multicolumn{4}{|c|}{MNIST} &\multicolumn{4}{|c|}{CIFAR-10} & \multicolumn{4}{|c}{SVHN} \\ \hline 
      Length (bit) & 12 & 24 & 32 & 48 & 12 & 24 & 32 & 48 & 12 & 24 & 32 & 48 \\ \hline \hline
      KSH \cite{Liu12} & 24.30 & 36.63 & 31.10 & 33.25 &17.65 & 14.80 & 15.50 & 16.63 & 24.18 & 24.36 & 24.72 & 21.87\\
      ITQ \cite{Gong13} & 37.63 & 53.87 & 51.76 & 54.11 & 12.93 & 14.06 & 13.40 & 15.11 & 16.22 & 16.85 & 19.67 & 19.89\\
      BRE \cite{Brian09} & 47.03 & 53.17 & 54.76 & 57.11 & 22.90 & 24.16 & 23.98 & 19.31 & 19.27 & 19.95 & 21.60 & 22.09\\
      DSH \cite{Liu16} & 96.05 & 97.35 & 98.10 & 98.13 & 38.17 & 38.70 & 40.19 & 37.29 & 73.16 & 70.33 & 82.16 & 77.33\\
      CNNH+ \cite{Rongkai14} & 97.57 & 97.89 & 98.04 & 98.33 & 40.00 & 42.00 & 44.89 & 44.55 & 78.32 & 81.46 & 81.81 & 84.00\\
      SSDH \cite{Yang18} & 98.83 & 98.97 & 98.96 & 99.15 & 82.31 & 84.07 & 83.78 & 84.28 & 93.19 & 93.98 & 93.95 & 94.46\\
      SFHC \cite{Hanjiang15} & 90.09 & 90.93 & 93.04 & 96.18 & 58.07 & 58.74 & 61.39 & 53.19 & 83.56 & 78.23 & 76.16 & 79.83\\
      DCAH \cite{Masoumeh18} & 99.50 & 99.44 & 99.53 & 99.54 & 85.85 & 87.38 & 87.00 & 86.33 & 95.48 & 96.28 & 95.95 & 96.45\\
      PDH & {\bf 99.73} & {\bf 99.74} & {\bf 99.78} & {\bf 99.77} & {\bf 95.26} & {\bf 95.81} & {\bf 96.04} & {\bf 95.58} & {\bf 96.90} & {\bf 96.90} & {\bf 97.04} & {\bf 97.06} \\ \hline
    \end{tabular}
    \label{tab:mAP}
  \end{center}
\normalsize
\vspace{-2pc}
\end{table*}
\begin{table}[t]
\footnotesize
  \begin{center}
    \caption{Precision@k of PDH.}
    \begin{tabular}{l|p{7mm}p{7mm}p{7mm}p{7mm}p{7mm}p{7mm}} \hline
      Datasets & $k$ \\ \hline 
       & 100 & 200 & 400 & 600 & 800 & 1000 \\ \hline \hline
      MNIST & 99.70 & 99.70 & 99.71& 99.70 & 99.70 & 99.71 \\
      CIFAR-10 & 95.03 & 95.07 & 95.12 & 95.14 & 95.15 & 95.15 \\
      SVHN & 96.85 & 96.90 & 96.94 & 96.95 & 96.96 & 96.96 \\ \hline
    \end{tabular}
    \label{tab:precision}
  \end{center}
  \normalsize
  \vspace{-2pc}
\end{table}
\vspace{-0.5pc}
\subsection{Image retrieval}
\label{subsec:retrieval}
\vspace{-0.5pc}
ANN search is performed using the hash function for query image $\bm{\tilde{p}}$ as shown in Fig.\ref{fig:architecture}.
%In a hashing method, it is necessary to convert an image to binary code and search by its Hamming distance.
The outputs of CNN $q_j(\bm{p_k})$ and $q_j(\bm{\tilde{p}})$ are continuous values, thus they have to be binarize for generating hash codes.
For discretizing $q_j(\bm{p_k}), q_j(\bm{\tilde{p}})$ and $d(\bm{p_k},\bm{\tilde{p}})$, we calculate MAP estimation $\hat{d}(h(\bm{p_k}),h(\bm{\tilde{p}}))$ of the distance function $d(h(\bm{p_k}),h(\bm{\tilde{p}}))$.
\vspace{-0.5pc}
\begin{eqnarray}
&\scalebox{0.80}{$\displaystyle \hat{d}(h(\bm{p_k}),h(\bm{\tilde{p}})) $} = \scalebox{0.9}{$\displaystyle \argmax_{d(h(\bm{p_k}),h(\bm{\tilde{p}}))} Pr(d(h(\bm{p_k}),h(\bm{\tilde{p}})) | \bm{p_k},\bm{\tilde{p}}) $} \nonumber \\
\hspace{-0.4pc}=& \scalebox{0.80}{$\displaystyle \sum\nolimits_{j=1}^{n} \argmax_{|h_j(\bm{p_k}) - h_j(\bm{\tilde{p}})|} Pr(|h_j(\bm{p_k}) - h_j(\bm{\tilde{p}})| | \bm{p_k},\bm{\tilde{p}}) = \sum\nolimits_{j=1}^{n} \sigma_j ,$} \label{eq:dhat}
\end{eqnarray}
where $\sigma_j$ is defined as follows:
\vspace{-0.5pc}
\begin{eqnarray}
&\scalebox{0.9}{$\displaystyle  \sigma_j =
\begin{cases}
1 & \text{if } Pr(|h_j(\bm{p_k}) - h_j(\bm{\tilde{p}})| = 1 | \bm{p_k},\bm{\tilde{p}}) \geq 0.5 .\\
0 & \text{otherwise.} 
\end{cases} $} \label{eq:sigma}
\end{eqnarray}
Here, the condition for $\sigma_j = 1$ is expressed as follows:
\vspace{-0.5pc}
\begin{eqnarray}
& \scalebox{0.9}{$\displaystyle Pr(|h_j(\bm{p_k}) - h_j(\bm{\tilde{p}})| = 1 | \bm{p_k},\bm{\tilde{p}}) \geq 0.5 $} .\nonumber \\
\iff & \scalebox{0.9}{$\displaystyle q_j(\bm{p_k}) (1-q_j(\bm{\tilde{p}})) + (1-q_j(\bm{p_k})) q_j(\bm{\tilde{p}}) \geq 0.5 .$} \nonumber \\
\iff & \scalebox{0.85}{$\displaystyle  \hat{h}_j(\bm{p_k}) (1 - \hat{h}_j(\bm{\tilde{p}})) + (1 - \hat{h}_j(\bm{p_k})) \hat{h}_j(\bm{\tilde{p}}) = 1 ,$}
\label{eq:condition}
\end{eqnarray}
where $\hat{h}_j(\bm{p_k})$ is calculated as follows:
\vspace{-0.5pc}
\begin{eqnarray}
\scalebox{0.8}{$\displaystyle \hat{\bm{h}}(\bm{p_k}) = \{\hat{h}_j(\bm{p_k})\}_{j=1}^{n}$}  \text{, where } 
\scalebox{0.8}{$\displaystyle \hat{h}_j(\bm{p_k}) = $}
\begin{cases}
\scalebox{0.8}{$\displaystyle 1 $} & \scalebox{0.9}{$\displaystyle \text{if $q_j(\bm{p_k}) \geq 0.5$} .$} \\
\scalebox{0.8}{$\displaystyle 0 $} & \scalebox{0.9}{$\displaystyle \text{otherwise.} $}
\end{cases} \label{eq:hath}
\end{eqnarray}
Based on Eq.(\ref{eq:dhat}), Eq.(\ref{eq:sigma}), Eq.(\ref{eq:condition}), Eq.(\ref{eq:hath}), $\hat{d}(\bm{p_k}, \bm{\tilde{p}})$ is calculated as follows:
\vspace{-0.5pc}
\begin{eqnarray}
\scalebox{0.9}{$\displaystyle \hat{d}(\bm{p_k},\bm{\tilde{p}}) = \sum\nolimits_{j=1}^{n} |\hat{h}_j(\bm{p_k}) - \hat{h}_j(\bm{\tilde{p}})| .$}
\end{eqnarray}
Therefore, the MAP estimation of the ideal hash distance is obtained as the Hamming distance between $q_j(\bm{p_k})$ thresholded by 0.5.
%binarizing $q_j(\bm{p_k})$ with "0.5" as a threshold and calculating the Hamming distance therebetween.
We use $\hat{d}(\bm{p_k},\bm{\tilde{p}})$ for image retrieval.

In this way, the distance function in the PDH is theoretically derived from the probability distribution on the image space.
The Hamming distance between the hash codes obtained by the PDH is equivalent to the MAP estimation of the ideal hash distance and realizes a probabilistically explainable method including binarization of the CNN outputs.
\vspace{-0.5pc}
\section{Experiment}
\label{sec:experiment}
\vspace{-0.5pc}
In order to confirm the effectiveness of the proposed method, we conduct evaluation experiments using three datasets.
\vspace{-0.5pc}
\subsection{Experimental setting}
\label{subsec:setting}
\vspace{-0.5pc}
We use three image datasets: MNIST \cite{Lecun98}, CIFAR-10 \cite{Krizhevsky09}, SVHN \cite{Yuval11} for evaluation.
We divide these datasets into training sets and test sets, and perform hash learning using training sets.
For image retrieval, training sets and test sets are used as gallery and query, respectively.
MNIST contains handwritten images of numbers (0-9) and consists of 60,000 training set (6,000 per class) and 10,000 test set (1,000 per class).
CIFAR-10 contains 10 class natural images such as airplane and cat.
In CIFAR-10, we randomly selected 50,000 images (5,000 per class) from 60,000 images as a training set and the remaining 10,000 as a test set (1,000 per class).
SVHN contains house numbers in Google Street View images and consists of 73,257 training set and 26,032 test set.
As a likelihood estimator, we use AlexNet\cite{Alex12} for MNIST and ResNet-50\cite{Kaiming16} for CIFAR-10, SVHN, respectively.
We modify only the final layers of these networks to specific dimensional fully connected layers.
These network weights are initialized by ImageNet \cite{Olga15} pre-trained models.
We use the Stochastic Gradient Descent \cite{Herbert85} as an optimizer for hash learning.
We apply data augmentation including rotation, shift, flip and random erasing \cite{Zhun17} to images in the training sets.
\vspace{-0.5pc}
\subsection{Evaluation protocols}
\label{subsec:protocol}
\vspace{-0.5pc}
As performance indicators, mean average precision (mAP) and precision@$k$ are used.
The mAP is the mean of average precision for the images of the same class in the gallery, see \cite{QingYuan18} for detailed calculation method.
Precision@$k$ is the percentage of true neighbors among the top $k$ data retrieved from the gallery by ANN search.
See \cite{QingYuan18} for detailed calculation method.
In order to confirm the effectiveness of the proposed method, we compare the accuracies with the conventional methods of KSH \cite{Liu12}, ITQ \cite{Gong13}, BRE \cite{Brian09}, DSH \cite{Liu16}, CNNH+\cite{Rongkai14}, SSDH \cite{Yang18}, SFHC \cite{Hanjiang15}, DCAH\cite{Masoumeh18}.
We refer to \cite{Masoumeh18} on the accuracy of these conventional methods.
\vspace{-0.5pc}
\subsection{Experimental Results}
\label{subsec:mnist}
\vspace{-0.5pc}
The results of hashing methods are shown in the Table \ref{tab:mAP}, \ref{tab:precision}.
As shown in Table \ref{tab:mAP}, our PDH outperforms the other state-of-the-art methods on mAP.
%As the bit length of hash code decreases, mAP of PDH decreases slightly.
Even when using a much short hash code of $12$ bits, high precision is maintained, and it is possible to search with high precision.
Furthermore, as shown in Table \ref{tab:precision}, the precision@$k$ of our PDH is almost unchanged for the value of $k$.
Therefore, even in the case of obtaining a small-scale similar image or obtaining a large-scale similar image, the proposed method works effectively.
\vspace{-0.5pc}
\section{Conclusion}
\label{sec:conclusion}
\vspace{-0.5pc}
In this paper, we propose probabilistic deep hashing (PDH) which generates probabilistically explainable hash codes.
In our PDH, single loss function with no hyperparameter is derived from image probability distributions.
By using this loss function, we can learn CNN-based hash functions.
Furthermore, we show the probabilistic optimality of Hamming distance between hash codes.
Based on this property, image retrieval with high precision is achieved.
We evaluate the performance of hashing methods and show that our PDH outperforms the conventional state-of-the-art hashing methods.

\bibliographystyle{IEEEbib}
\bibliography{ms}

\end{document}